\documentclass[12pt]{article}	

\usepackage{amsmath,amsthm,amsfonts,amscd} 
\usepackage{verbatim}      	
\usepackage{makeidx}       	
\usepackage{epsfig}         	
\usepackage{url}		
				
\usepackage{graphicx}
\usepackage[caption=false]{subfig}
\graphicspath{{Figures/}} 
\usepackage{amsfonts}
\usepackage{algorithm}
\usepackage{algorithmicx}
\usepackage{algpseudocode}
\usepackage{tikz}
\usepackage{float}
\usepackage{xcolor}
\usepackage{dsfont}

\newtheorem{theorem}{Theorem}[section]

\newtheorem{lemma}[theorem]{Lemma}

\theoremstyle{definition}
\newtheorem{definition}{Definition}[section]

\theoremstyle{remark}

\newcommand\restr[2]{{
  \left.\kern-\nulldelimiterspace 
  #1 
  \vphantom{\big|} 
  \right|_{#2} 
  }}

\newcommand{\latexe}{{\LaTeX\kern.125em2%
                      \lower.5ex\hbox{$\varepsilon$}}}

\chardef\bslash=`\\	

\makeatletter		
\def\square{\RIfM@\bgroup\else$\bgroup\aftergroup$\fi
  \vcenter{\hrule\hbox{\vrule\@height.6em\kern.6em\vrule}%
                                              \hrule}\egroup}
\makeatother		

\makeindex    

\usepackage{amsmath,mathtools}

\newcommand{\beq}{\begin{equation} \begin{aligned}}
\newcommand{\eeq}{\end{aligned} \end{equation}}
\newcommand{\beqstar}{\begin{equation*} \begin{aligned}}
\newcommand{\eeqstar}{\end{aligned} \end{equation*}}

\newcommand{\proj}{\mathcal{P}}

\newcommand{\map}{\phi} 
\newcommand{\vis}{\psi} 
\newcommand{\graze}{\gamma} %

\newcommand{\free }{\Omega_{\text{free}}} 
\newcommand{\obs }{\Omega_{\text{obs}}} 
\newcommand{\seqset}{\mathbf{A}}
\newcommand{\vanpts }{O} 
\newcommand{\visarea }{f} 
\newcommand{\visset }{\mathcal{V}} 
\newcommand{\exgap}{\rho}

\usepackage[mathscr]{eucal}
\newcommand{\envsubset}{\mathscr{S}}
\newcommand{\envset}{\mathcal{S}}
\newcommand{\cumvisset}{\mathscr{V}}

\newcommand{\bigo}{\mathcal{O}}
\usepackage{upgreek}
\newcommand{\bigomega}{\Upomega}

\usepackage{graphicx}
\usepackage{comment}
\usepackage{amsmath,amssymb} 
\usepackage{color}

\usepackage{subfig}
\usepackage{xcolor}
\usepackage{paralist}
\usepackage[normalem]{ulem}   

\usepackage{siunitx}

\usepackage{dblfloatfix} 
\usepackage{balance}  

\newcommand{\Colin}[1]{{\textcolor{orange}{ \textbf{cbm:} #1 }}}

\renewcommand{\Colin}[1]{}

\usepackage[doublespacing]{setspace}

\title{{\bf \Large{Greedy Algorithms for Sparse Sensor Placement via Deep Learning \vspace{1em} }}}

\author{Louis Ly and Yen-Hsi Richard Tsai\\
\footnotesize{\tt{louisly@utexas.edu, ytsai@math.utexas.edu} }\\[3em]
Oden Institute for Computational Engineering and Sciences \\ 
The University of Texas at Austin} 
\date{}

\begin{document}
\maketitle

\begin{abstract}
We consider the exploration problem: an agent equipped with a depth sensor must map out a previously unknown environment using as few sensor measurements as possible. We propose an approach based on supervised learning of a greedy algorithm.  We provide a bound on the optimality of the greedy algorithm using submodularity theory.  Using a level set representation, we train a convolutional neural network to determine vantage points that maximize visibility. We show that this method drastically reduces the on-line computational cost and determines a small set of vantage points that solve the problem. This enables us to efficiently produce highly-resolved and topologically accurate maps of complex 3D environments. Unlike traditional next-best-view and frontier-based strategies, the proposed method accounts for geometric priors while evaluating potential vantage points. While existing deep learning approaches focus on obstacle avoidance and local navigation, our method aims at finding near-optimal solutions to the more global exploration problem. We present realistic simulations on 2D and 3D urban environments.
\end{abstract}

\section{Introduction}

We consider the problem of generating a minimal sequence of observing locations
to achieve complete line-of-sight visibility coverage of an environment.  In
particular, we are interested in the case when environment is initially
unknown.  This is particularly useful for autonomous agents to map out unknown,
or otherwise unreachable environments, such as undersea caverns.  Military
personnel may avoid dangerous situations by sending autonomous agents to scout
new territory.  We first assume the environment is known in order to gain
insights.

Consider a domain $\Omega \subseteq \mathbb{R}^d$.  Partition the domain
$\Omega=\free\cup\obs$ into an open set $\free$ representing the free space,
and a closed set $\obs$ of finite obstacles without holes. We will refer to the
$\obs$ as the environment, since it is characterized by the obstacles.
Let $x_i\in\free$ be a vantage point, from which a range sensor,
such as LiDAR, takes omnidirectional measurements $\proj_{x_i}:S^{d-1}\to\mathbb{R}$.
That is, $\proj_{x_i}$ outputs the distance to closest obstacle for each direction in the unit sphere.
One can map the range measurements to the visibility set $\visset_{x_i}$;
points in $\visset_{x_i}$ are visible from $x_i$:

\beq
x \in \visset_{x_i} \text{ if } \|x-x_i\|_2 < \proj_{x_i} \Big( \frac{x-x_i}{\|x-x_i\|_2} \Big) 
\eeq

As more range
measurements are acquired, $\free$ 
can be approximated by the \emph{cumulatively visible set} $\Omega_k$:
\beq \Omega_k = \bigcup_{i=0}^k \mathcal{V}_{x_i} \eeq
By construction, $\Omega_k$ admits partial ordering: $\Omega_{i-1} \subset \Omega_{i}$.
For suitable choices of $x_i$, it is possible that
$ \Omega_n \to \free$
(say, in the Hausdorff distance).

We aim at determining a \emph{minimal set of vantage points} $\vanpts$ from which
every $x \in \free$ can be seen.
One may formulate a constrained optimization problem and
look for sparse solutions. When the environment is known, we have the \emph{surveillance} problem:
\beq\label{eq:surveillance-problem}
\min_{\vanpts\subseteq \free} \ | \vanpts | ~~~\text{subject to }  \free=\bigcup_{x\in \vanpts} \visset_{x} \ .
\eeq

When the environment is not known apriori, the agent must be careful to avoid collision with obstacles.
New vantage points must be a point that is currently visible. That is, $x_{k+1} \in \Omega_k$.
Define the set of admissible sequences:
\beq
\seqset(\free):=\{ (x_0,\dots,x_{n-1}) \ |\ n\in\mathbb{N},\ x_0\in\free, \ x_{k+1}\in\Omega_k\}.
\eeq
For the unknown environment, we have the \emph{exploration} problem:
\beq\label{eq:exploration-problem}
\min_{\vanpts \in\seqset(\free)} \ |\vanpts| ~~~\text{subject to }  \free=\bigcup_{x\in \vanpts} \visset_{x}.
\eeq
The problem is feasible as long as obstacles do not have holes.

   \begin{figure}[hptb]
      \centering
      \includegraphics[width=2in]{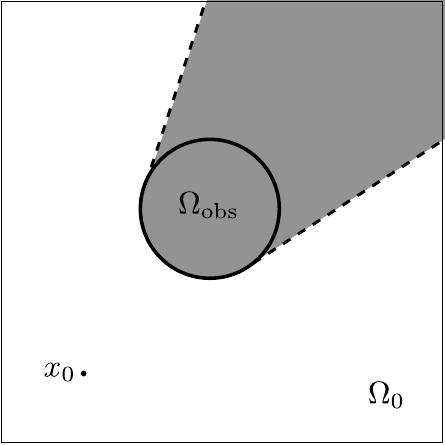} \quad
      \includegraphics[width=2in]{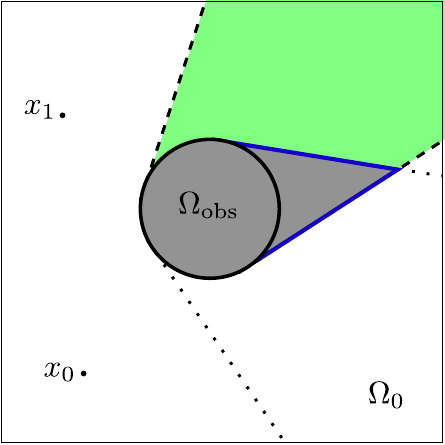}
      \caption{An illustration of the environment. Dashed and dotted lines are the
horizons from $x_0$ and $x_1$, respectively. Their shadow boundary, $B_1$,
is shown in thick, solid blue.  The area of the green region represents
$g(x_1; \Omega_0)$.  }

      \label{fig:setup}
   \end{figure}

\subsection{Related works}

The surveillance problem is related to the art gallery problem in computational
geometry, where the task is to determine the minimum set of guards who can
together observe a polygonal gallery. Vertex guards must be stationed at the
vertices of the polygon, while point guards can be anywhere in
the interior. For simply-connected polygonal scenes, Chv\'atal
showed that $\lfloor n/3 \rfloor$ vertex guards, where $n$ is the number of vertices,
are sometimes necessary and always sufficient \cite{chvatal1975combinatorial}.
For polygonal scenes with $h$ holes, $\lfloor (n+h)/3 \rfloor$ point
guards are sufficient \cite{bjorling1995efficient,hoffmann1991art}. However,
determining the optimal set of observers is NP-complete
\cite{urrutia2000art,o1983some,lee1986computational}.


Goroshin et al. propose an alternating minimization scheme for optimizing the
visibility of $N$ observers \cite{goroshin2011approximate}.
Kang et al. use a system of differential
equations to optimize the location and orientation of $N$ sensors to maximize
surveillance \cite{kang2017optimal}. Both works assume the number of sensors is given.

For the exploration problem, the ``wall-following'' strategy may be used to map
out simple environments \cite{zhang2004experimental}.  LaValle and
Tovar et al.  \cite{tovar2004gap,lavalle2006planning,tovar2007distance} combine
wall-following with a gap navigation tree to keep track of gaps, critical
events which hide a connected region of the environment that is occluded from a
vantage point. Exploration is complete when all gaps have been eliminated.
This approach does not produce any geometric representation of the environment upon completion, due to limited information from gap sensors.

A class of approaches pick new
vantage points along shadow boundaries (aka frontiers), the boundary between free
and occluded regions \cite{yamauchi1997frontier}.
Ghosh et al. propose a frontier-based approach for 2D polygonal environments which requires $r+1$ views,
where $r$ is the number of reflex angles \cite{ghosh2008online}.
For general 2D environments, Landa et al.
\cite{landa2006visibility,landa2007robotic,landa2008visibility} use high order
ENO interpolation to estimate curvature, which is then used to determine how
far past the horizon to step. 
However, it is not necessarily optimal to pick only points along the shadow
boundary, e.g. when the map is a star-shaped polygon \cite{ghosh2008online}.

Next-best-view algorithms try to find vantage points that maximize a utility
function, consisting of some notion of \emph{information gain} and another
criteria such as path length.  The vantage point does not have to lie along the
shadow boundary.  A common measure of information gain is the volume of \emph{entire}
unexplored region within sensor range that is not occluded by obstacles
\cite{gonzalez2002navigation, bircher2016receding, bircher2018receding,
heng2015efficient}. Surmann et al. count the number of intersections of rays
into the occlusion \cite{surmann2003autonomous}, while Valente et
al. \cite{valente2014information} use
the surface area of the shadow boundary, weighted by the viewing angle from the vantage points,
to define potential information gain. The issue with these heurisitics is
that they are independent of the underlying geometry. 
In addition, computing the information gain at each potential vantage point is
costly and another heurisitic is used to determine which points to sample.

There has been some attempts to incorporate deep learning into the
exploration problem, but they focus on navigation rather than
exploration.
The approach of Bai et al. \cite{bai2017toward} terminates when there is no
occlusion within view of the agent, even if the global map is still incomplete.
Tai and Liu \cite{tai2016mobile,tai2017virtual,lei2016robot} train agents to
learn obstacle avoidance. 

Our work uses a gain function to steer a greedy approach, similar to the
next-best-view algorithms. However, our measure of information gain takes the
geometry of the environment into account. By taking advantage of precomputation
via convolutional neural networks, our model learns shape priors for a large
class of obstacles and is efficient at runtime. We use a volumetric
representation which can handle arbitrary geometries in 2D and 3D. Also, we
assume that the sensor range is larger than the domain, which makes the problem
more global and challenging.


\section{Greedy algorithm}
\label{sec:greedy}
%
%

We propose a greedy approach which sequentially determines a new vantage point, $x_{k+1}$, based on the information gathered from all previous vantage points, $x_0,x_1,\cdots, x_{k}$.
The strategy is greedy because $x_{k+1}$ would be a location that \emph{maximizes the information gain}.

For the surveillance problem, the environment is known. We define the \emph{gain} function:
\begin{equation}g(x;\Omega_k) := |   \visset_x  \cup \Omega_k| - |\Omega_k |, \label{gain-func} \end{equation}
i.e. the volume of the region that is visible from $x$ but not from $x_0,x_1,\cdots,x_{k}$.
Note that $g$ depends on $\obs$, which we omit for clarity of notation.
The next vantage point should be chosen to maximize the newly-surveyed volume. We define the greedy surveillance algorithm as:
\begin{equation} x_{k+1} = \arg \max_{x\in \free} g(x;\Omega_k).\label{eq:greedy-surv}\end{equation}

The problem of exploration is even more challenging since, by definition, the
environment is not known. Subsequent vantage points must lie within the current visible set $\Omega_k$.
The corresponding greedy exploration algorithm is
\begin{equation} x_{k+1} = \arg \max_{x\in \Omega_k} g(x;\Omega_k).\label{eq:greedy-exp}\end{equation}

However, we remark that in practice, one is
typically interested only in a subset $\envsubset$ of all possible environments
$\envset:=\{\obs|\obs\subseteq\mathbb{R}^d\}$.

For example, cities generally follow a grid-like pattern.  Knowing these priors
can help guide our estimate of $g$ for certain types of $\obs$, even when
$\obs$ is unknown initially.

We propose to encode these priors formally into the parameters, $\theta$, of a learned function:
\beq g_\theta(x; \Omega_k, B_k ) \text{ for } \obs \in \envsubset, \eeq
where $B_k$ is the part of $\partial{\Omega}_k$ that may actually lie in the free space $\free$:
\beq
B_k &= \partial \Omega_k \backslash \obs.
\eeq

\begin{figure}[hptb]
      \vspace{.8em}
      \centering
      \includegraphics[height=1.4in,trim={0 0.11in 0 0.11in},clip]{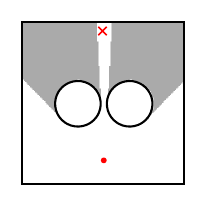}
      \includegraphics[height=1.4in,trim={0 0.11in 0 0.11in},clip]{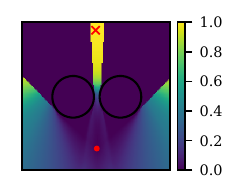}
	  \caption{ Left: the map of a scene consisting of two disks. Right: the
                intensity of the corresponding gain function. The current vantage point is
                shown as the red dot. The location which maximizes the gain function is shown
                as the red {\tt x}.} \label{fig:gain} 
	  \end{figure}

See Figure \ref{fig:gain} for an example gain function.
We shall demonstrate that while training for $g_\theta$, incorporating the shadow boundaries 
helps, in some sense, localize the learning of $g$, and is essential in creating usable $g_\theta$.

\subsection{A bound for the known environment}
\label{sec:bound-surv}
We present a bound on the optimality of the greedy algorithm, based on
submodularity \cite{krause2014submodular}, a useful property of set functions.
We start with standard definitions.
Let $V$ be a finite set and $f:2^V \to \mathbb{R}$ be a set function which
assigns a value to each subset $S\subseteq V$. 

\begin{definition}{(Monotonicity)}
A set function $f$ is \emph{monotone} if for
every $A\subseteq B \subseteq V$, $$f(A)\le f(B).$$
\end{definition}

\begin{definition}{(Discrete derivative)}
The \emph{discrete derivative} of $f$ at $S$ with respect to $v\in V$
is $$\Delta_f(v|S):= f(S\cup \{v\}) - f(S).$$
\end{definition}

\begin{definition}{(Submodularity)} A set function $f$ is \emph{submodular} if
for every $A\subseteq B\subseteq V$ and $v\in V\setminus B$, 
$$\Delta_f(v|A) \ge \Delta_f(v|B).$$
\end{definition}

In other words, set functions are submodular if they have diminishing returns.
More details and extensions of submodularity can be found in \cite{krause2014submodular}.

Now, suppose the environment $\obs$ is known.
Let $\vanpts$ be the set of vantage points, and let $\visarea(\vanpts)$ be the 
volume of the region visible from $\vanpts$:
\beq
\cumvisset(\vanpts) &:= \bigcup_{x\in\vanpts} \visset_x \\
\visarea(\vanpts) &:= \Big| \cumvisset(\vanpts) \Big|
\eeq

\begin{lemma}{} 
The function $\visarea$ is monotone.
\end{lemma}

\begin{proof}
Consider $A\subseteq B\subseteq \free$. Since $\visarea$ is the cardinality of unions of sets, we have
\beqstar
\visarea(B) &= \Big| \bigcup_{x\in B} \visset_x  \ \Big| \\ 
 &= \Big| \bigcup_{x\in A\cup\{B\setminus A \}} \visset_x  \ \Big| \\ 
 &\ge \Big| \bigcup_{x\in A} \visset_x  \ \Big| \\ 
 &= \visarea(A).
\eeqstar

\end{proof}

\begin{lemma}{} 
The function $\visarea$ is submodular.
\end{lemma}

\begin{proof}
Suppose $A\subseteq B$ and $\{v\} \in \free\setminus B$. By properties of unions and intersections, we have
\beqstar
\visarea(A\cup \{v\}) + \visarea(B) &= \Big| \bigcup_{x\in (A\cup \{v\})} \visset_x  \Big| + \Big| \bigcup_{x\in B} \visset_x  \Big| \\
&\ge \Big| \bigcup_{x\in A\cup \{v\}\cup B} \visset_x  \Big| + \Big| \bigcup_{x\in (A\cup\{v\}) \cap B} \visset_x  \Big| \\
&= \Big| \bigcup_{x\in  B\cup \{v\}} \visset_x  \Big| + \Big| \bigcup_{x\in A} \visset_x  \Big| \\
&= \visarea( B\cup \{v\}) + \visarea(A)\\
\eeqstar
Rearranging, we have
\beqstar
\visarea(A\cup \{v\}) + \visarea(B) &\ge \visarea( B\cup \{v\}) + \visarea(A)\\
\visarea(A\cup \{v\}) - \visarea(A) &\ge \visarea( B\cup \{v\}) - \visarea(B)\\
\Delta_\visarea (v|A) &\ge \Delta_\visarea(v|B).
\eeqstar%
\end{proof}
Submodularity and monotonicity enable a bound which compares the relative
performance of the greedy algorithm to the optimal solution.


\begin{theorem}{}
\label{thm:bound-surv}
Let $\vanpts_k^\ast$ be the optimal set of $k$ sensors.
Let $\vanpts_n= \{x_i\}_{i=1}^n$ be the set of $n$ sensors placed using the greedy surveillance algorithm \eqref{eq:greedy-surv}.
Then,
$$\visarea(\vanpts_n) \ge (1-e^{-n/k}) \visarea(\vanpts_k^\ast) .$$
\end{theorem}

\begin{proof}

For $l<n$ we have    
\begin{align}
\visarea(\vanpts_k^\ast) &\le \visarea(\vanpts_k^\ast \cup \vanpts_l) \label{eq:bound-mono}\\ 
&= \visarea(\vanpts_l) + \Delta_\visarea(\vanpts_k^\ast | \vanpts_l) \\
&= \visarea(\vanpts_l) +  \sum_{i=1}^k \Delta_\visarea(x_i^\ast|\vanpts_l \cup\{x_1^\ast,\dots,x_{i-1}^\ast\} ) \\
&\le \visarea(\vanpts_l) +  \sum_{i=1}^k \Delta_\visarea(x_i^\ast|\vanpts_l) \label{eq:bound-sub}\\
&\le \visarea(\vanpts_l) +  \sum_{i=1}^k \visarea(\vanpts_{l+1}) - \visarea(\vanpts_l) \label{eq:bound-greedy} \\
&= \visarea(\vanpts_l)   + k \big[ \visarea(\vanpts_{l+1}) - \visarea(\vanpts_l)  \big].
\end{align}
Line (\ref{eq:bound-mono}) follows from monotonicity, (\ref{eq:bound-sub}) follows from submodularity
of $\visarea$, and (\ref{eq:bound-greedy}) from definition of the greedy algorithm.
   Define $\delta_l := \visarea(\vanpts_k^\ast) - \visarea(\vanpts_l)$, with $\delta_0:=\visarea(\vanpts_k^\ast)$. 
Then 
   \beqstar
   \visarea(\vanpts_k^\ast) - \visarea(\vanpts_l) &\le k \big[ \visarea(\vanpts_{l+1}) - \visarea(\vanpts_l)\big] \\
   \delta_l &\le k \big[ \delta_l - \delta_{l+1}\big] \\
   \delta_l \Big( 1 -  k \Big) &\le - k \delta_{l+1} \\
   \delta_l \Big( 1 -  \frac{1}{k} \Big) &\ge \delta_{l+1} \\
   \eeqstar
  Expanding the recurrence relation with $\delta_n$, we have 
   \beqstar
   \delta_n &\le  \Big( 1 -  \frac{1}{k} \Big) \delta_{n-1}\\
   &\le \Big( 1-\frac{1}{k} \Big)^n \delta_0 \\
   &= \Big( 1-\frac{1}{k} \Big)^n \visarea(\vanpts_k^\ast) \\
   \eeqstar

	Finally, substituting back the definition for $\delta_n$, we have the desired result:
    \begin{align}
    \delta_n \le  \Big( 1-\frac{1}{k} \Big)^n \visarea(\vanpts_k^\ast) \nonumber \\
    \visarea(\vanpts_k^\ast) - \visarea(\vanpts_n) \le  \Big( 1-\frac{1}{k} \Big)^n \visarea(\vanpts_k^\ast) \nonumber \\
    \visarea(\vanpts_k^\ast)\Big( 1- (1-1/k)^n\Big) \le \visarea(\vanpts_n) \nonumber\\
    \visarea(\vanpts_k^\ast)\Big( 1- e^{-n/k} \Big) \le \visarea(\vanpts_n) \label{eq:bound-exp}
    \end{align}
  where (\ref{eq:bound-exp}) follows from the inequality $1-x\le e^{-x}$.
\end{proof}

In particular, if $n=k$, then $(1-e^{-1})\approx 0.63$. 
This means that $k$ steps of the greedy algorithm is guaranteed to cover at least 63\%
of the total volume, if the optimal solution can also be obtained with $k$ steps. When $n=3k$, the greedy algorithm covers at least 95\% of the total volume.
In \cite{nemhauser1978best}, it was shown that no polynomial time algorithm can achieve a better bound.

\subsection{A bound for the unknown environment}
\label{sec:bound-exp}
When the environment is not known, subsequent vantage points must lie within
the current visible set to avoid collision with obstacles:
\beq
x_{k+1} \in \cumvisset(\vanpts_k)
\eeq
Thus, the performance of the exploration algorithm has a strong dependence on the environment
$\obs$ and the initial vantage point $x_1$. We characterize this dependence
using the notion of the \emph{exploration ratio}.

Given an environment $\obs$ and $A\subseteq\free$, consider the ratio of the marginal value of the greedy exploration algorithm, 
to that of the greedy surveillance algorithm:
\beq
\exgap(A) &:= \frac{ \displaystyle{\sup_{x\in \cumvisset(A)}}\Delta_\visarea(x|A ) }{ \displaystyle{ \sup_{x\in\free}} \Delta_\visarea(x|A)}.
\eeq
That is, $\exgap(A)$ characterizes the relative gap (for lack of a better word) caused by the collision-avoidance constraint $x\in\cumvisset(A)$.
Let $A_x =\{A\subseteq \free|x\in A\}$ be the set of vantage points which contain $x$. 
Define the \emph{exploration ratio} as
\beq
\exgap_x &:= \inf_{A\in A_x} \exgap(A).
\eeq

The exploration ratio is the worst-case gap between the two greedy
algorithms, conditioned on $x$.
It helps to provide a bound for the difference between the optimal
solution set of size $k$, and the one prescribed by $n$ steps the greedy exploration algorithm.

\begin{theorem}{}
\label{thm:bound-exp}
Let $\vanpts_k^\ast =\{x_i^\ast\}_{i=1}^k$ be the optimal sequence of $k$ sensors which includes $x_1^\ast=x_1$.
Let $\vanpts_n= \{x_i\}_{i=1}^n$ be the sequence of $n$ sensors placed using the greedy exploration algorithm
\eqref{eq:greedy-exp}.
Then, for $k,n>1$:
$$\visarea(\vanpts_n) \ge \Big[1-\exp \Big({\frac{-(n-1)\exgap_{x_1}}{k-1}}\Big) \Big(1-\frac{\visarea(x_1)}{\visarea(\vanpts_k^\ast)} \Big)    \Big] \visarea(\vanpts_k^\ast) .$$
\end{theorem}

This is reminiscent of Theorem~\ref{thm:bound-surv}, with two subtle differences. 
The $\big[1-\frac{\visarea(x_1)}{\visarea(\vanpts_k^\ast)}\big]$ term accounts for the shared vantage point $x_1$.
If $\visarea(x_1)$ is large, then the exponential term has little effect, since $\visarea(x_1)$ is already close to $\visarea(\vanpts_k^\ast)$. On the other hand, if it is small, then
the exploration ratio $\exgap_{x_1}$ plays a factor.
The idea of the proof is similar, with some subtle differences in algebra to account for the shared vantage point
$x_1$, and the exploration ratio $\exgap_{x_1}$.

\begin{proof}

We have, for $l<n$:
\begin{align}
\visarea(\vanpts_k^\ast) &\le \visarea(\vanpts_k^\ast \cup \vanpts_l) \nonumber\\ 
&= \visarea(\vanpts_l) + \Delta_\visarea(\vanpts_k^\ast | \vanpts_l) \nonumber\\
&= \visarea(\vanpts_l) +  \sum_{i=1}^k \Delta_\visarea(x_i^\ast|\vanpts_l \cup\{x_1^\ast,\dots,x_{i-1}^\ast\} ) \label{eq:bd-exp-tele}\\
&\le \visarea(\vanpts_l) +  \sum_{i=1}^k \Delta_\visarea(x_i^\ast|\vanpts_l) \label{eq:bd-exp-sub}\\
&= \visarea(\vanpts_l) + \Delta_\visarea(x_1^\ast|\vanpts_l) +  \sum_{i=2}^k \Delta_\visarea(x_i^\ast|\vanpts_l) \nonumber\\
&= \visarea(\vanpts_l) + \sum_{i=2}^k \Delta_\visarea(x_i^\ast|\vanpts_l) \label{eq:bd-exp-x1}\\
&\le \visarea(\vanpts_l) +  \sum_{i=2}^k \max_{x\in\free} \Delta_\visarea(x|\vanpts_l) \nonumber\\
&\le \visarea(\vanpts_l) +  \frac{1}{\exgap_{x_1}} \sum_{i=2}^k \max_{x\in\cumvisset(\vanpts_l)} \Delta_\visarea(x|\vanpts_l) \label{eq:bd-exp-gap}\\
&\le \visarea(\vanpts_l) +  \frac{1}{\exgap_{x_1}} \sum_{i=2}^k \visarea(\vanpts_{l+1}) - \visarea(\vanpts_l) \label{eq:bd-exp-greedy} \\
&= \visarea(\vanpts_l)   + \frac{k-1}{\exgap_{x_1}} \big[ \visarea(\vanpts_{l+1}) - \visarea(\vanpts_l)  \big]. \nonumber
\end{align}
Line \eqref{eq:bd-exp-tele} is a telescoping sum, \eqref{eq:bd-exp-sub} follows from submodularity of $\visarea$,
\eqref{eq:bd-exp-x1} uses the fact that $x_1^\ast\in\vanpts_l$,
\eqref{eq:bd-exp-gap} follows from the definition of $\exgap_{x_1}$ and \eqref{eq:bd-exp-greedy} stems from the definition
of the greedy exploration algorithm \eqref{eq:greedy-exp}.

As before, define $\delta_l := \visarea(\vanpts_k^\ast) - \visarea(\vanpts_l)$. However, this time, note
that $\delta_1:=\visarea(\vanpts_k^\ast) - \visarea(\vanpts_1) = \visarea(\vanpts_k^\ast) - \visarea(x_1) $. 
Then 
   \beqstar
   \visarea(\vanpts_k^\ast) - \visarea(\vanpts_l) &\le \frac{k-1}{\exgap_{x_1}} \big[ \visarea(\vanpts_{l+1}) - \visarea(\vanpts_l)\big] \\
   \delta_l &\le \frac{k-1}{\exgap_{x_1}} \big[ \delta_l - \delta_{l+1}\big] \\
   \delta_l \Big( 1 -  \frac{k-1}{\exgap_{x_1}}  \Big) &\le -  \frac{k-1}{\exgap_{x_1}} \delta_{l+1} \\
   \delta_l \Big( 1 -  \frac{\exgap_{x_1}}{k-1} \Big) &\ge \delta_{l+1} \\
   \eeqstar
  Expanding the recurrence relation with $\delta_n$, we have 
   \beqstar
   \delta_n &\le  \Big( 1 -  \frac{\exgap_{x_1}}{k-1} \Big) \delta_{n-1}\\
   &\le \Big( 1-\frac{\exgap_{x_1}}{k-1} \Big)^{n-1} \delta_1 \\
   &= \Big( 1-  \frac{\exgap_{x_1}}{k-1} \Big)^{n-1}  \big[ \visarea(\vanpts_k^\ast) -\visarea(x_1)\big] \\
   \eeqstar
	Now, substituting back the definition for $\delta_n$, we arrive at
    \begin{align}
    \delta_n &\le  \Big( 1-\frac{\exgap_{x_1}}{k-1} \Big)^{n-1} \big[ \visarea(\vanpts_k^\ast)-\visarea(x_1)\big] \nonumber \\
    \visarea(\vanpts_k^\ast) - \visarea(\vanpts_n) &\le  \Big( 1-\frac{\exgap_{x_1}}{k-1} \Big)^{n-1} \big[ \visarea(\vanpts_k^\ast)-\visarea(x_1)\big] \nonumber \\
    \visarea(\vanpts_k^\ast) -\visarea(x_1) -\big[ \visarea(\vanpts_n)-\visarea(x_1)\big]  &\le  \Big( 1-\frac{\exgap_{x_1}}{k-1} \Big)^{n-1} \big[ \visarea(\vanpts_k^\ast)-\visarea(x_1)\big] \nonumber \\
    \big[\visarea(\vanpts_k^\ast) -\visarea(x_1)\big] \Big(1- \big[ 1-\frac{\exgap_{x_1}}{k-1} \big]^{n-1}\Big)  &\le   \big[ \visarea(\vanpts_n)-\visarea(x_1)\big] \nonumber \\
    \big[\visarea(\vanpts_k^\ast) -\visarea(x_1)\big] \Big(1- e^{-\frac{(n-1)\exgap_{x_1}}{k-1}} \Big)  &\le   \big[ \visarea(\vanpts_n)-\visarea(x_1)\big] \nonumber.
    \end{align}

Finally, with some more algebra
\begin{align}
\big[ \visarea(\vanpts_n)-\visarea(x_1)\big] &\ge    \Big(1- e^{-\frac{(n-1)\exgap_{x_1}}{k-1}} \Big) \big[\visarea(\vanpts_k^\ast) -\visarea(x_1)\big] \nonumber \\
\visarea(\vanpts_n) &\ge  \visarea(x_1)  +  \Big(1- e^{-\frac{(n-1)\exgap_{x_1}}{k-1}} \Big) \big[\visarea(\vanpts_k^\ast) -\visarea(x_1)\big] \nonumber \\
\visarea(\vanpts_n) &\ge  \visarea(x_1)  +  \Big(1- e^{-\frac{(n-1)\exgap_{x_1}}{k-1}} \Big) \visarea(\vanpts_k^\ast) - \visarea(x_1) + \visarea(x_1)   e^{-\frac{(n-1)\exgap_{x_1}}{k-1}}   \nonumber \\
\visarea(\vanpts_n) &\ge    \Big(1- e^{-\frac{(n-1)\exgap_{x_1}}{k-1}} \Big) \visarea(\vanpts_k^\ast) + \visarea(x_1)  e^{-\frac{(n-1)\exgap_{x_1}}{k-1}}  \nonumber \\
\visarea(\vanpts_n) &\ge    \Big(1- e^{-\frac{(n-1)\exgap_{x_1}}{k-1}} \big[ 1-\frac{\visarea(x_1)}{\visarea(\vanpts_n^\ast)} \big] \Big) \visarea(\vanpts_k^\ast)  \nonumber.
\end{align}

\end{proof}

\subsubsection*{Exploration ratio example}
\begin{figure}[hptb]
      \centering
      \includegraphics[width=.5\textwidth]{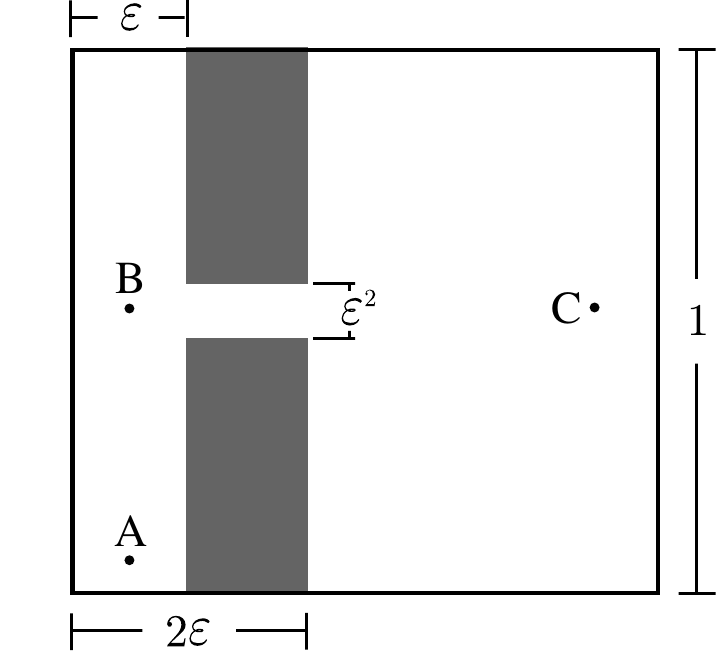}
	  \caption{A map with a narrow alley. Scale exaggerated for illustration.}
		\label{fig:alley} 
	\end{figure}

We demonstrate an example where $\exgap_{x}$ can be an arbitrarily small
factor that is determined by the geometry of $\free$.
Figure~\ref{fig:alley} depicts an illustration of the setup for the narrow alley environment.

Consider a domain $\Omega=[0,1]\times[0,1]$ with a thin vertical wall of width $\varepsilon\ll1$,
whose center stretches from $(\frac{3}{2}\varepsilon,0)$ to
$(\frac{3}{2}\varepsilon,1)$. A narrow opening of size
$\varepsilon^2\times\varepsilon$ is centered at
$(\frac{3}{2}\varepsilon,\frac{1}{2})$.  
Suppose $x_1=x_1^\ast=A$ so that 
$$\visarea(\{x_1\})=\varepsilon +\bigo(\varepsilon^2),$$
where the $\varepsilon^2$ factor is due to the small sliver of the narrow alley visible from $A$.
By observation, the optimal solution contains two vantage points. One such solution places $x_2^\ast =C$.
The greedy exploration algorithm can only place $x_2\in\cumvisset(x_1)=[0,\varepsilon]\times[0,1]$.
One possible location is $x_2=B$.
Then, after 2 steps of the greedy algorithm, we have $$\visarea(\vanpts_2)=\varepsilon+\bigo(\varepsilon^2).$$
Meanwhile, the total visible area is $$\visarea(\vanpts_2^\ast)=1-\bigo(\varepsilon)$$ 
and the ratio of greedy to optimal area coverage is 
\beq
\label{eq:alley-greedy-ratio-1}
\frac{\visarea(\vanpts_2)}{\visarea(\vanpts_2^\ast)} =\frac{\varepsilon+\bigo(\varepsilon^2)}{1-\bigo(\varepsilon)} =  \bigo(\varepsilon)
\eeq
The exploration ratio is $\exgap_{x_1}=\bigo(\varepsilon^2)$, since
\beq
\max_{x\in\cumvisset(\{x_1\})} \Delta_\visarea(x|\{x_1\})&=\bigo(\varepsilon^2) \\
\max_{x\in\free} \Delta_\visarea(x|\{x_1\})&=1-\bigo(\varepsilon) \\
\eeq
According to the bound, with $k=n=2$, we should have 
\beq
\frac{\visarea(\vanpts_2)}{\visarea(\vanpts_2^\ast)} &\ge  \Big(1- e^{-\frac{(n-1)\exgap_{x_1}}{k-1}} \big[ 1-\frac{\visarea(x_1)}{\visarea(\vanpts_2^\ast)} \big] \Big) \\
&= \Big(1- e^{-\bigo(\varepsilon^2)} \big[ 1-\bigo(\varepsilon) \big] \Big)\\
&= \bigomega(\varepsilon)
\eeq
which reflects what we see in \eqref{eq:alley-greedy-ratio-1}.

On the other hand, if $\vanpts_2=\{C,B\}$ and $\vanpts_2^\ast=\{C,B\}$, we would have
$$\visarea(\{x_1\})=1-\bigo(\varepsilon)$$ and $\exgap_{x_1}=1$,
since both the greedy exploration and surveillance step coincide.
According to the bound, with $k=n=2$, we should have 
\beq
\frac{\visarea(\vanpts_2)}{\visarea(\vanpts_2^\ast)} &\ge  \Big(1- e^{-\frac{(n-1)\exgap_{x_1}}{k-1}} \big[ 1-\frac{\visarea(x_1)}{\visarea(\vanpts_n^\ast)} \big] \Big) \\
&\ge 1-\bigo(\varepsilon)
\eeq
which is the case, since $\visarea(\vanpts_2)=\visarea(\vanpts_2^\ast)$.

By considering the first vantage point $x_1$ as part of the bound,
we account for some of the unavoidable uncertainties associated with unknown environments during exploration.


\subsection{Numerical comparison}
\label{sec:bound-num}
We compare both greedy algorithms on random arrangements of up to 6 circular obstacles. 
Each algorithm starts from the same initial position and runs until all free area is covered.
We record the number of vantage points required over 200 runs for each number of obstacles.

\begin{figure}[hptb]
      \centering
      \includegraphics[width=.99\textwidth]{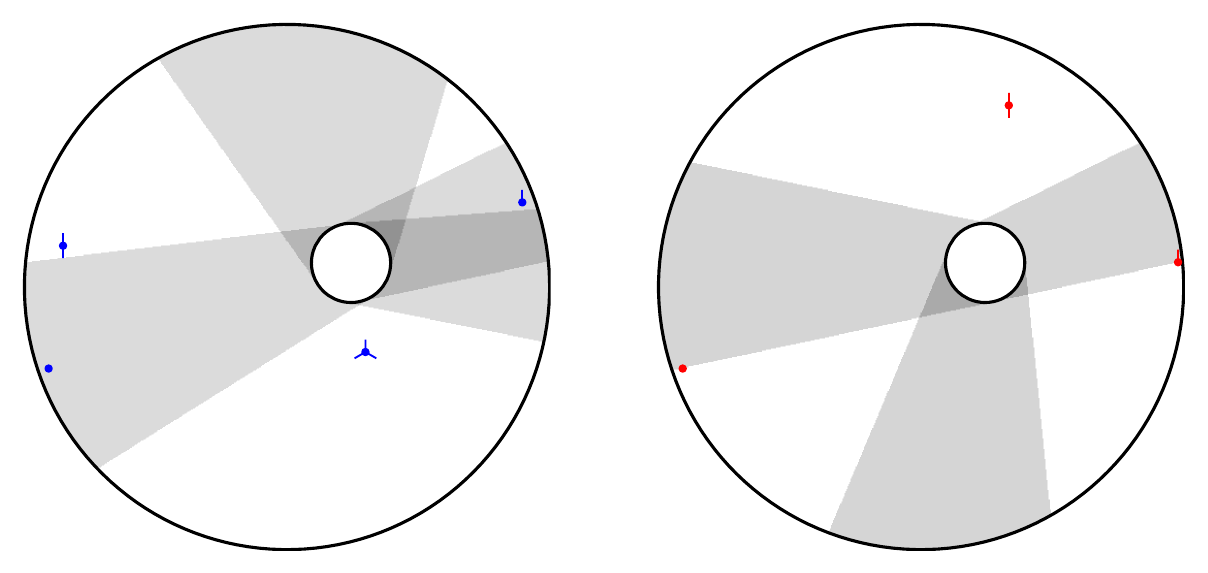}
      \includegraphics[width=.99\textwidth]{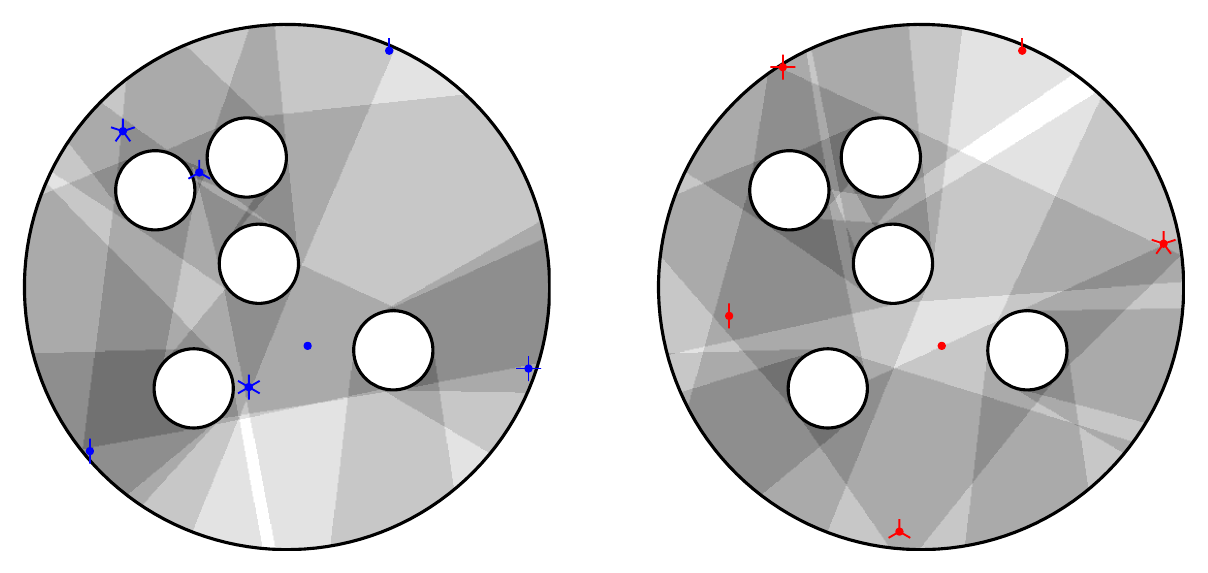}
	  \caption{Comparing the greedy algorithm for the known (left) and unknown (right) environment on circular obstacles.
               Spikes on each vantage point indicate the ordering, e.g. the initial point has no spike. Gray areas
               are shadows from each vantage point. Lighter regions are visible from more vantage points.}
		\label{fig:circles-135} 
	\end{figure}

\begin{figure}[hptb]
      \centering
      \includegraphics[height=.93\textheight]{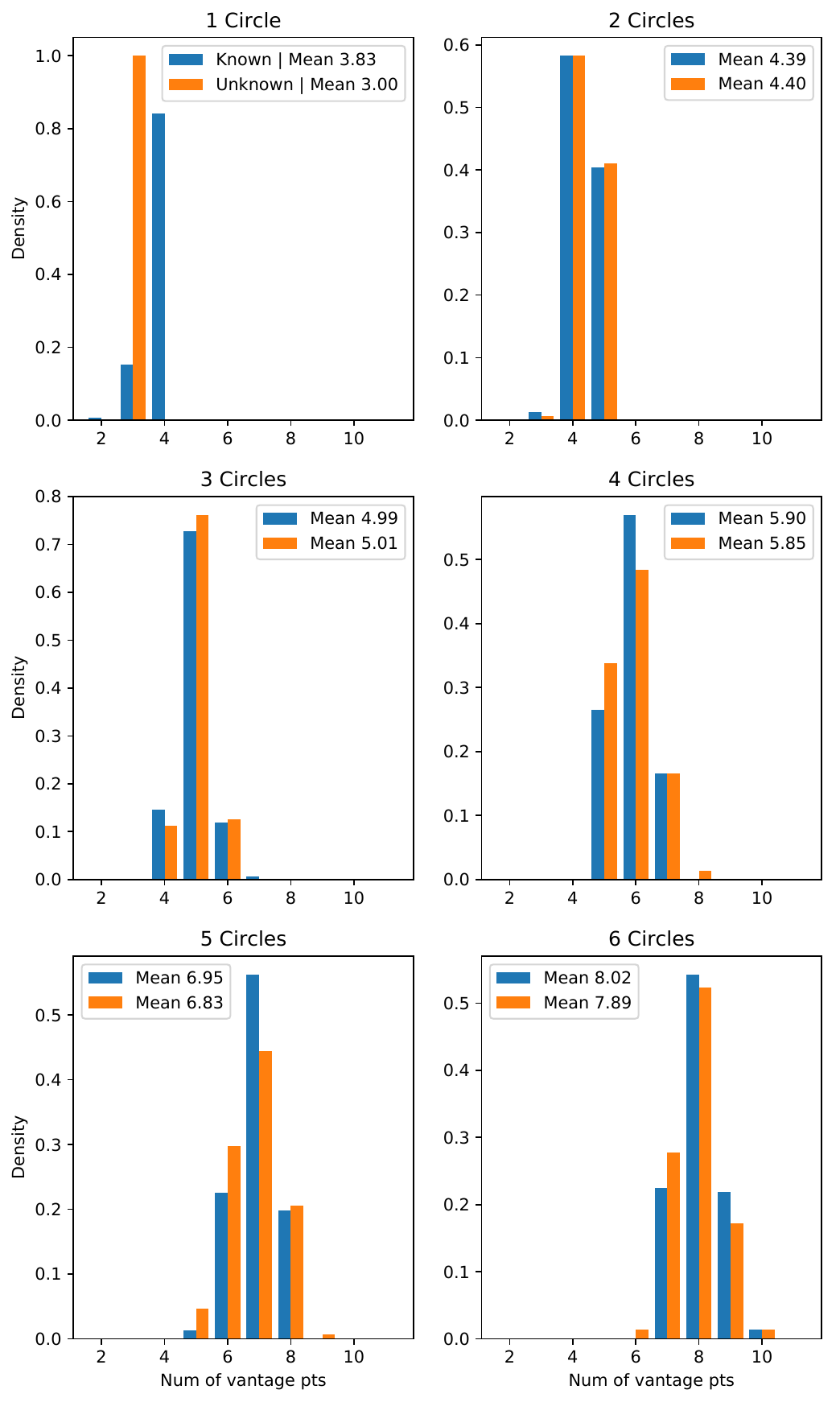}
	  \caption{Histogram of number of vantage points needed for the
surveillance (blue) and exploration (orange) greedy algorithms to completely
cover environments consisting up of to 6 circles.}
		\label{fig:circles-histogram} 
	\end{figure}

Surprisingly, the exploration algorithm sometimes requires fewer vantage points
than the surveillance algorithm. Perhaps the latter is too aggressive, or perhaps the collision-avoidance
constraint acts as a regularizer.
For example, when there is a single circle, the greedy surveillance algorithm places the
second vantage point $x_2$ on the opposite side of this obstacle. This may lead to two slivers
of occlusion forming of either side of the circle, which will require 2 additional vantage points to cover.
With the greedy exploration algorithm, we do not have this problem, due to the collision-avoidance constraint.
Figure~\ref{fig:circles-135} shows an select example with 1 and 5 obstacles. 
Figure~\ref{fig:circles-histogram} show the histogram of the number of steps needed
for each algorithm. On average, both algorithms require a similar number of steps, but
the exploration algorithm has a slight advantage.


\section{Learning the gain function}
In this section, we discuss the method for approximating the gain function when the map is not known.
Given the set of previously-visited vantage points, we compute the cumulative visibility
and shadow boundaries. We approximate the gain function by applying the trained
neural network on this pair of inputs, and pick the next point according to
\eqref{eq:greedy-surv}. This procedure repeats until there are no shadow
boundaries or occlusions. 

The data needed for the training and evaluation of $g_\theta$ are computed
using level sets \cite{osher1988fronts,sethian1999level,osher2006level}.
Occupancy grids may be applicable, but we choose level sets since they have proven
to be accurate and robust. In particular, level sets are necessary for subpixel
resolution of shadow boundaries and they allow for efficient visibility
computation, which is crucial when generating the library of training
examples.

The training geometry is embedded by a signed distance function, denoted by $\map$.
For each vantage point $x_i$, the visibility set is represented by the level
set function $\vis(\cdot,x_i)$, which is computed efficiently using the algorithm described in
\cite{tsai2004visibility}.

In the calculus of level set functions, unions and intersections of sets are
translated, respectively, into taking maximum and minimum of the corresponding characteristic
functions.  The cumulatively visible sets  $\Omega_k$  are  represented by  the
level set function $ \Psi_k(x)$, which is defined recursively by
\begin{align}
\Psi_0(x) &=\vis(x,{x_0}), \\
\Psi_k(x) &=\max \left\{ \Psi_{k-1}(x), \vis(x,{x_k}) \right\}, \quad k=1,2,\dots
\end{align}
where the max is taken point-wise. Thus we have 
\begin{align}
\free &= \{x|\map(x) >0\},\\
\visset_{x_i} &= \{x|\vis(x,{x_i})>0\},\\
\Omega_k &= \{x|\Psi_k(x) >0\}.
\end{align}
The shadow boundaries $B_k$
are approximated by the "smeared out" function:
\begin{equation} 
b_k(x) := \delta_\varepsilon (\Psi_k) \cdot \left[ 1-H(G_k(x))  \right],
\end{equation}
where $H(x)$ is the Heaviside function and 
\begin{align}
\delta_\varepsilon(x) &= \frac{2}{\varepsilon}  \cos^2\left( \frac{\pi x }{\varepsilon} \right) \cdot \mathds{1}_{[-\frac{\varepsilon}{2},\frac{\varepsilon}{2}]} (x), \\
\graze(x,x_0) &= (x_0-x)^T \cdot \nabla \map(x),\\
G_0 &= \graze(x,x_0),\\
G_k(x) &= \max\{G_{k-1}(x),\graze(x,x_k)\}, \quad k=1,2,\dots
\end{align}

Recall, the shadow boundaries are the portion of the $\partial \Omega_k$ that lie in free space;
the role of $1-H(G_k)$ is to mask out the portion of obstacles that are currently visible from
$\{x_i\}_{i=1}^k$. See Figure~\ref{fig:ls_notation} for an example of $\graze$.
In our implementation, we take
$\varepsilon=3\Delta x$ where $\Delta x$ is the grid node spacing. We refer
the readers to \cite{tsai2005total} for a short review of  relevant
details.

When the environment $\obs$ is known, we can compute the gain function exactly
\begin{equation}g(x; \Omega_k) = \int H \Big( H\big(\vis(\xi,x)\big) - H\big( \Psi_k(\xi) \big)\Big) \ d\xi. \end{equation}
We remark that the integrand will 
be 1 where the new vantage point uncovers something not previously seen.
Computing $g$ for all $x$ is costly; 
each visibility and volume computation requires $\bigo(m^d)$ operations, and repeating this for all points
in the domain results in $\bigo(m^{2d})$ total flops. 
We approximate it with a function ${\tilde g}_\theta$ parameterized
by $\theta$: 
\beq {\tilde g}_\theta(x; \Psi_k, \map, b_k) \approx g(x; \Omega_k).
\eeq

If the environment is unknown, we directly approximate the gain function
by learning the parameters $\theta$ of a function 
\beq
g_\theta(x;\Psi_k,b_k) \approx  g(x; \Omega_k) H(\Psi_k) 
\eeq
using only the observations as input. Note the $H(\Psi_k)$ factor is needed for
collision avoidance during exploration because it is not known \emph{a priori} whether
an occluded location $y$ is part of an obstacle or free space.  Thus $g_\theta(y)$ must be zero.

\subsection{Training procedure}

We sample the environments uniformly from a library.  For each $\obs$, a sequence of
data pairs is generated and included into the training set $\mathcal{T}$:
\beq \big(\{ \Psi_k, b_k\}, g(x;\Omega_k)H(\Psi_k) \big), \qquad k=0,1,2,\dots.\eeq

\begin{figure}[ht]
      \centering
      \includegraphics[width=.98\textwidth]{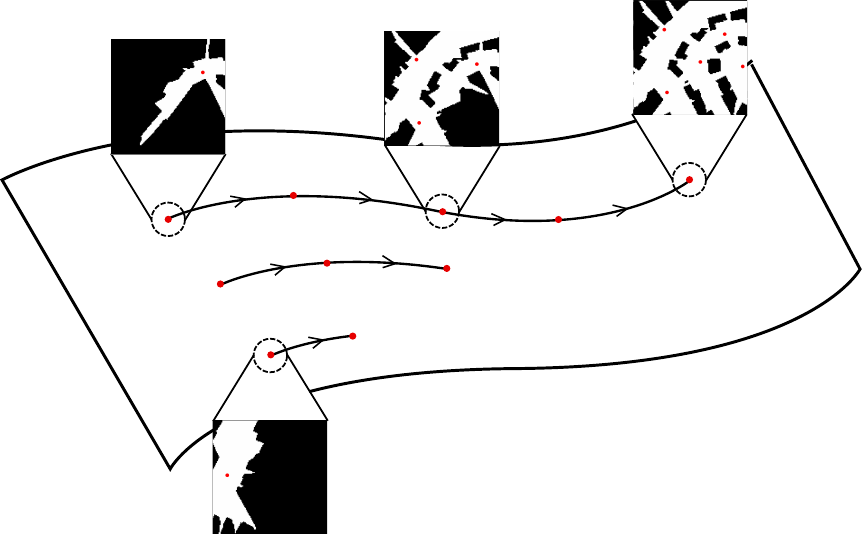}
	  \caption{Causal data generation along the subspace of relevant shapes.
               Each dot is a data sample corresponding to a sequence of vantage points.}
		\label{fig:manifold} 
	\end{figure}

For a given environment $\obs$, define a path $\vanpts=\{x_i\}_{i=0}^k$
as admissible if $\map(x_0)>0$ and $\Psi_i(x_{i+1})>0$ for $i=0,\dots,k-1$.
That is, it should only contain points in free space and
in the case of exploration, subsequent points must be visible from at least one
of the previous vantage points.
Let $\mathcal{A}$ be the set of admissible paths.
Then training set should ideally include all paths in $\mathcal{A}$. However
this is too costly, since there are $\bigo(m^{kd})$ paths consisting of $k$ steps.
Instead, to generate causally relevant data, we use an $\varepsilon$-greedy
approach: we uniformly sample initial positions. With probability
$\varepsilon$, the next vantage point is chosen randomly from admissible set.
With probability $1-\varepsilon$, the next vantage point is chosen according to
\eqref{eq:greedy-surv}. Figure~\ref{fig:manifold} shows an illustration of the
generation of causal data along the subspace of relevant shapes.

The function $g_\theta$ is
learned by minimizing the empirical loss across all data pairs for each $\obs$
in the training set $\mathcal{T}$:
\beq
\underset{\theta}{\mathrm{argmin}} \ \frac{1}{N}\sum_{\obs \in \mathcal{T} } \sum_k L\Big(g_\theta(x;\Psi_k,b_k), g(x;\Omega_k) H(\Psi_k) \Big),
\eeq
where $N$ is the total number of data pairs. We use the cross entropy loss function:
\beq
 L(p,q)=\int p(x) \log q(x) + (1-p(x)) \log (1-q(x)) \ dx.
\eeq

    \begin{figure}[hptb]
       \centering
       \subfloat[][a)]{\includegraphics[height=.25\textheight,trim={0 0.02in 0 0.02in},clip]{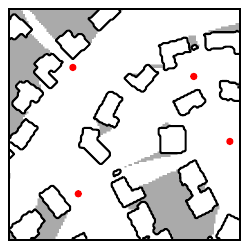}}\quad
       \subfloat[][b)]{\includegraphics[height=.25\textheight,trim={0 0.02in 0 0.02in},clip]{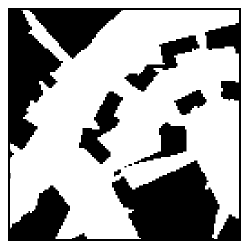}}\phantom{test}\\
       \subfloat[][c)]{\includegraphics[height=.25\textheight,trim={0.01in 0.02in 0 0.02in},clip]{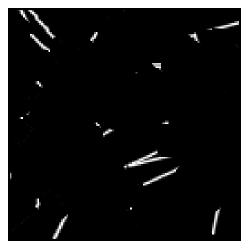}}\quad
       \subfloat[][d)]{\includegraphics[height=.25\textheight,trim={0 0.02in 0 0.02in},clip]{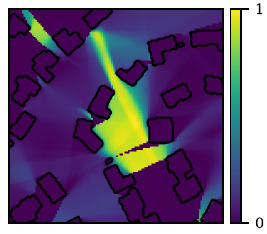}}\\

	   \caption{A training data pair consists of the cumulative visibility and
		shadow boundaries as input, and the gain function as the output. Each sequence of vantage
        points generates a data sample which depends strongly the shapes of the obstacles and shadows. a) The
		underlying map with current vantage points shown in red. b) The cumulative
		visibility of the current vantage points. c) The corresponding shadow
		boundaries. d) The corresponding gain function.}
	   \label{fig:training-data} \end{figure}

\subsection*{Network architecture}

We use convolutional neural networks (CNNs) to approximate the gain function,
which depends on the shape of $\obs$ and the location $x$. CNNs have been
used to approximate functions of shapes effectively in many applications.
Their feedforward evaluations are efficient if the off-line training cost is
ignored.  The gain function $g(x)$ does not depend \emph{directly} on $x$, but
rather, $x$'s visibility of $\free$, with a domain of dependence bounded by
the sensor range. 
We employ a fully convolutional
approach for learning $g$, which makes the network applicable to domains of
different sizes. The generalization to 3D is also straight-forward.

We base the architecture of the CNN on U-Net \cite{ronneberger2015u}, which has
had great success in dense inference problems, such as image segmentation. It
aggregates information from various layers in order to have wide receptive
fields while maintaining pixel precision.  The main design choice is to make
sure that the receptive field of our model is sufficient. That is, we want to
make sure that the value predicted at each voxel depends on a sufficiently
large neighborhood.  For efficiency, we use convolution kernels of size $3$ in
each dimension.  By stacking multiple layers, we can achieve large receptive
fields.  Thus the complexity for feedforward computations is linear in the
total number of grid points. 



Define a \emph{conv block} as the following layers:
convolution, batch norm, leaky {\tt relu},
stride 2 convolution, batch norm, and leaky {\tt relu}.
Each \emph{conv block} reduces the image size by a factor of 2.
The latter half of the network increases the image size using \emph{deconv blocks}:
bilinear 2x upsampling, convolution, batch norm, and leaky {\tt relu}.

Our 2D network uses 6 \emph{conv blocks} followed by 6 \emph{deconv blocks},
while our 3D network uses 5 of each block.  We choose the number of blocks to
ensure that the receptive field is at least the size of the training images: $128\times128$
and $64\times64\times64$.  The first \emph{conv block} outputs 4 channels.
The number of channels doubles with each \emph{conv block}, and halves with
each \emph{deconv block}.

The network ends with a single channel, kernel of size~1 convolution layer
followed by the sigmoid activation.  This ensures that the network aggregates all
information into a prediction of the correct size and range.


\section{Numerical results}

We present some experiments to demonstrate the efficacy of our approach. Also, we demonstrate its limitations.
First, we train on $128\times128$ aerial city blocks cropped from INRIA Aerial Image Labeling Dataset \cite{maggiori2017dataset}.
It contains binary images with building labels from several urban areas, including Austin, Chicago, Vienna, and Tyrol.
We train on all the areas except Austin, which we hold out for evaluation.
We call this model {\bf City-CNN}. We train a similar model {\bf NoSB-CNN} on the same training data, but omit the shadow boundary from the input.
Third, we train another model {\bf Radial-CNN}, on synthetically-generated radial maps, such
as the one in Figure \ref{fig:shapes-gain}.

Given a map, we randomly select an initial location.
In order to generate the sequence of vantage points, we apply
\eqref{eq:greedy-surv}, using $g_\theta$ in place of $g$.  Ties are broken by
choosing the closest point to $x_k$.  We repeat this process until there are no
shadow boundaries, the gain function is smaller than $\epsilon$, or the
residual is less than $\delta$,  where the residual is defined as:
\begin{equation} \label{eq:residual}
r = \frac{|\free \setminus \Omega_k| } {|\free| } .
\end{equation}

We compare these against the algorithm which uses the exact gain function,
which we call {\bf Exact}.  We also compare against {\bf Random}, a random
walker, which chooses subsequent vantage points uniformly from the visible
region, and {\bf Random-SB} which samples points uniformly in a small neighborhood
of the shadow boundaries. We analyze the number of steps required to cover the scene and the
residual as a function of the number of steps.
   \begin{figure}[htbp]
      \vspace{.8em}
      \centering
      \includegraphics[height=1.8in]{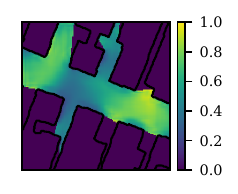}
      \includegraphics[height=1.8in]{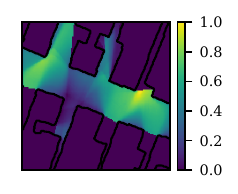}

	  \caption{Comparison of predicted (left) and exact (right) gain function
		for an Austin map. Although the functions are not identical, the predicted gain
		function peaks in similar locations to the exact gain function, leading to
		similar steps.}
		\label{fig:gain-compare} 
	\end{figure}

Lastly, we present simulation for exploring 3D environments.
Due to the limited availability of datasets, the model, {\bf 3D-CNN}, is
trained using synthetic $64\times64\times64$ voxel images consisting of
tetrahedrons, cylinders, ellipsoids, and cuboids of random positions, sizes,
and orientations.
In the site\footnote{http://visibility.page.link/demo}, the
interested reader may inspect the performance of the {\bf 3D-CNN} in some other
challenging 3D environments.


For our experiments using trained networks, we make use of a CPU-only machine containing
four Intel Core i5-7600 CPU @ 3.50GHz and 8 GB of RAM.
Additionally, we use an Nvidia Tesla K40 GPU with 12 GB of memory for training
and predicting the gain function in 3D scenes.

   \begin{figure}[htb]
      \vspace{.8em}
      \centering
      \includegraphics[height=1.3in]{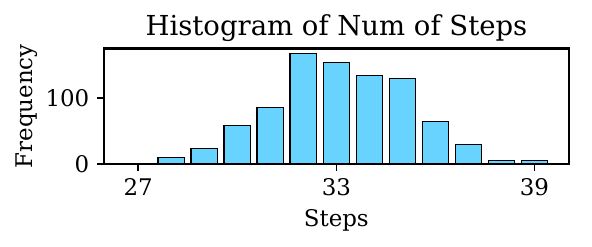}
      \includegraphics[height=1.3in]{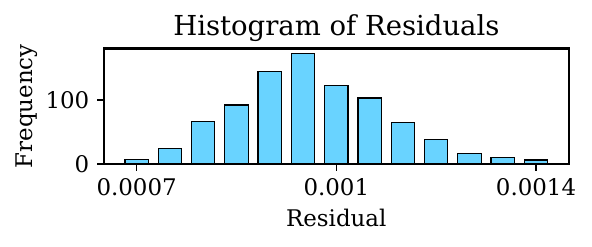}
      \caption{Distribution of the residual and number of steps generated
across multiple runs over an Austin map. The proposed method is robust against
varying initial conditions. The algorithm reduces the residual to roughly 0.1 \% within
39 steps by using a threshold on the predicted gain function as a termination
condition. }

      \label{fig:city_stats}
   \end{figure}

   \begin{figure}[htpb]
      \centering
      \includegraphics[height=4in,trim={0 0.02in 0 0.02in},clip]{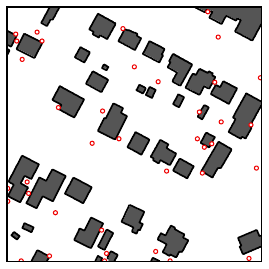}
      \caption{\label{fig:austin} An example of 36 vantage points (red disks) using {\bf City-CNN} model. White regions are free space while gray regions are occluded. Black borders indicate edges of obstacles.}
   \end{figure}

   \begin{figure}[htb]
      \centering
      \includegraphics[width=4in]{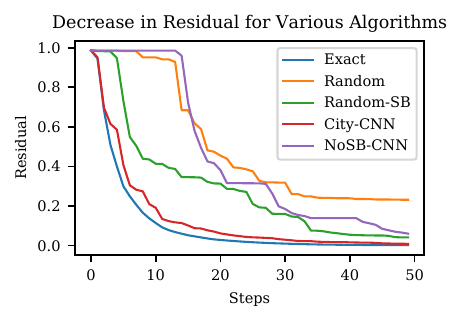}

      \caption{Graph showing the decrease in residual over 50 steps among
	   various algorithms starting from the same initial position for an Austin map.
	   Without using shadow boundary information, {\bf NoSB-CNN} can at times be worse than
	   {\bf Random}. Our {\bf City-CNN} model is significantly faster than {\bf Exact}
	   while remaining comparable in terms of residual.}
      \label{fig:comparison}
   \end{figure}

\subsection*{2D city}
The {\bf City-CNN} model works well on 2D Austin maps.
First, we compare the predicted gain function to the exact gain function on a $128\times128$ map, as in Figure \ref{fig:gain-compare}.
Without knowing the underlying map, it is difficult to accurately determine the gain function.
Still, the predicted gain function peaks in locations similar to those in the exact gain function. This results
in similar sequences of vantage points.

\emph{The algorithm is robust to the initial positions.} Figure~\ref{fig:city_stats}
show the distribution of the number of steps and residual across over 800 runs
from varying initial positions over a $512\times512$ Austin map. In practice, using the
shadow boundaries as a stopping criteria can be unreliable.  Due to numerical
precision and discretization effects, the shadow boundaries may never
completely disappear.  Instead, the algorithm terminates when the maximum
predicted gain falls below a certain threshold $\epsilon$. In this example, we
used $\epsilon = 0.1$. Empirically, this strategy is robust. On average, the
algorithm required 33 vantage points to reduce the occluded region to within
0.1\% of the explorable area.

Figure \ref{fig:austin} shows an example sequence consisting of 36 vantage points.
Each subsequent step is generated in under 1 sec using the CPU and instantaneously with a GPU.

Even when the maximizer of the predicted gain function is different from that
of the exact gain function, the difference in gain is negligible.  This is
evident when we see the residuals for {\bf City-CNN} decrease at similar rates
to {\bf Exact}.  Figure \ref{fig:comparison} demonstrates an example of the
residual as a function of the number of steps for one such sequence generated
by these algorithms on a $1024\times1024$ map of Austin.
We see that {\bf City-CNN} performs comparably to {\bf Exact} approach in terms
of residual.  However, {\bf City-CNN} takes 140 secs to generate 50 steps on
the CPU while {\bf Exact}, an $\bigo(m^4)$ algorithm, takes more than 16
hours to produce 50 steps.



   \begin{figure}[htpb]
	\vspace{.5em}
      \centering
      \includegraphics[height=4in,trim={0 0.02in 0 0.02in},clip]{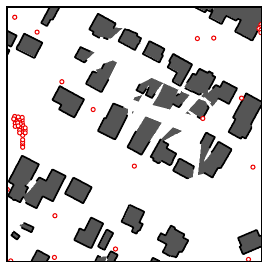}
      \caption{\label{fig:nosb} A sequence of 50 vantage points generated from {\bf NoSB-CNN}. The points cluster near flat edges due to ambiguity and the algorithm becomes stuck. Gray regions without black borders have not been fully explored.}
   \end{figure}

\subsection*{Effect of shadow boundaries}
\emph{The inclusion of the shadow boundaries as input to the CNN is critical for the algorithm to work.}
Without the shadow boundaries, the algorithm cannot distinguish between obstacles and occluded regions.
If an edge corresponds to an occluded region, then choosing a nearby vantage point will 
reduce the residual. However, choosing a vantage point near a flat obstacle will result in no change to the
cumulative visibility. At the next iteration, the input is same as the previous iteration, and the result will be the same;
the algorithm becomes stuck in a cycle. To avoid this, we prevent vantage points from repeating
by zeroing out the gain function at that point and recomputing the argmax.
Still, the vantage points
tend to cluster near flat edges, as in Figure \ref{fig:nosb}. This clustering behavior causes the {\bf NoSB-CNN} model to
be, at times, worse than {\bf Random}. See Figure \ref{fig:comparison} to see how the clustering inhibits the reduction
in the residual.

\subsection*{Effect of shape}

\begin{figure}[phtb]
	\vspace{2em}
       \centering
       \subfloat[][a)]{\includegraphics[height=1.5in,trim={0 0.11in 0 0.11in},clip]{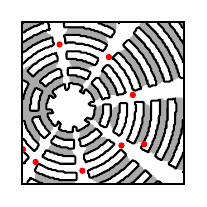}}\phantom{filler}\quad
       \subfloat[][b)]{\includegraphics[height=1.5in,trim={0 0.11in 0 0.11in},clip]{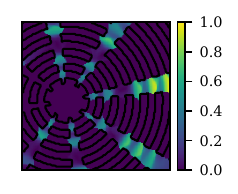}}\\
       \subfloat[][c)]{\includegraphics[height=1.5in,trim={0 0.11in 0 0.11in},clip]{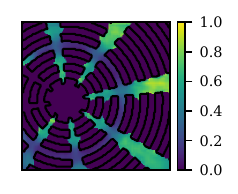}}\quad
       \subfloat[][d)]{\includegraphics[height=1.5in,trim={0 0.11in 0 0.11in},clip]{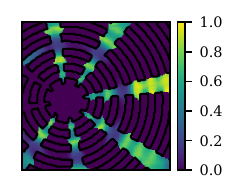}}\\

       \caption{Comparison of gain functions produced with various models on a
radial scene. Naturally, the CNN model trained on radial obstacles best
approximates the true gain function. a) The underlying radial map with vantage
points show in red. b) The exact gain function c)  {\bf City-CNN} predicted
gain function. d) {\bf Radial-CNN} predicted gain function.}

       \label{fig:shapes-gain}
    \end{figure}

The shape of the obstacles, i.e. $\Omega^c$, used in training affects the gain function predictions.
Figure \ref{fig:shapes-gain} compares the gain functions produced by {\bf City-CNN} and {\bf Radial-CNN}.

   \begin{figure}[htb]
      \centering
      \includegraphics[height=4in,trim={0 0.02in 0 0.02in},clip]{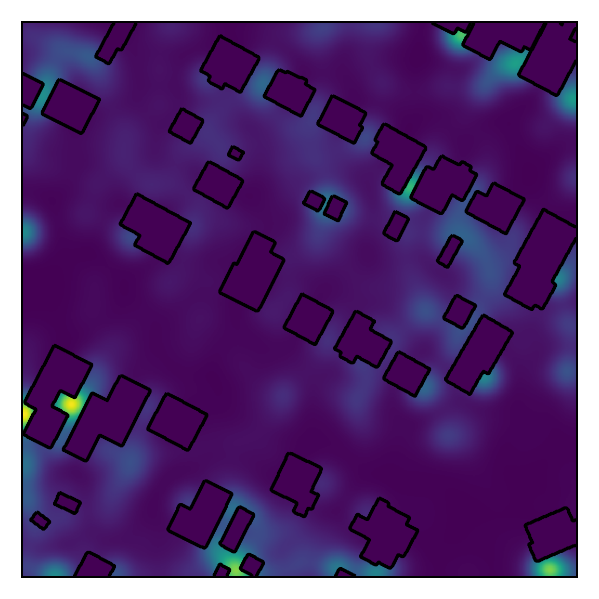}
      \caption{\label{fig:frequency} Distribution of vantage points generated by {\bf City-CNN} method from various initial positions. Hot spots are brighter and are visited more frequently since they are essential for completing coverage.}
   \end{figure}

\subsection*{Frequency map}

Here we present one of our studies concerning the exclusivity of
vantage point placements in $\Omega$. We generated sequences of vantage points
starting from over 800 different initial conditions using {\bf City-CNN} model
on a $512\times512$ Austin map. Then, we model each vantage point as a Gaussian with
fixed width, and overlay the resulting distribution on the Austin map in Figure
\ref{fig:frequency}.  This gives us a frequency map of the most recurring
vantage points. These hot spots reveal regions that are more secluded and
therefore, the visibility of those regions is more sensitive to vantage point
selection.
The efficiency of the CNN method
allows us to address many surveillance related
questions for a large collection of relevant geometries.

\subsection*{Art gallery}
Our proposed approach outperforms the computational geometry solution
\cite{o1987art} to the art gallery problem, even though we do not assume the
environment is known.  The key issue with computational geometry approaches is
that they are heavily dependent on the triangulation.
In an extreme  example, consider an art gallery that is a simple convex $\emph{n-gon}$. Even though it is sufficient to place a single vantage point anywhere in the interior of the room, the triangulation-based approach produces a solution with $\lfloor n/3 \rfloor$ vertex guards.

\begin{figure}[bpt]
      \centering
      \includegraphics[height=2.5in]{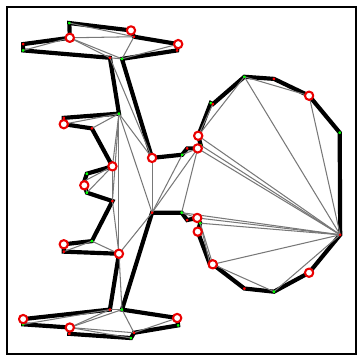}
      \includegraphics[height=2.5in]{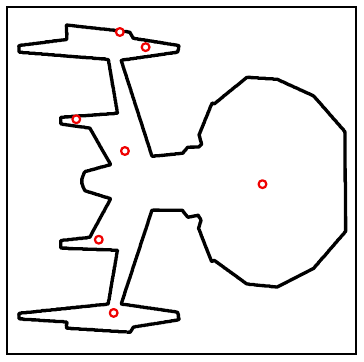}
      \caption{Comparison of the computational geometry approach and the {\bf City-CNN}
approach to the art gallery problem.  The red circles are the vantage points computed by the methods.
Left: A result computed by the computational geometry approach, given the environment.
Right: An example sequence of 7 vantage points generated by the {\bf City-CNN}
model. }

      \label{fig:art-gallery}
   \end{figure}

Figure \ref{fig:art-gallery} shows an example gallery consisting of 58 vertices. The computational
geometry approach requires $\lfloor\frac{n}{3}\rfloor=19$ vantage points to completely cover the scene,
even if point guards are used \cite{bjorling1995efficient,hoffmann1991art}.
The gallery contains $r=19$ reflex angles, so the work of \cite{ghosh2008online} requires $r+1=20$ vantage points.
On average, 
{\bf City-CNN}
requires only 8 vantage points.

\subsection*{3D environment}

\begin{figure}[htpb]
      \centering
      \includegraphics[width=5in]{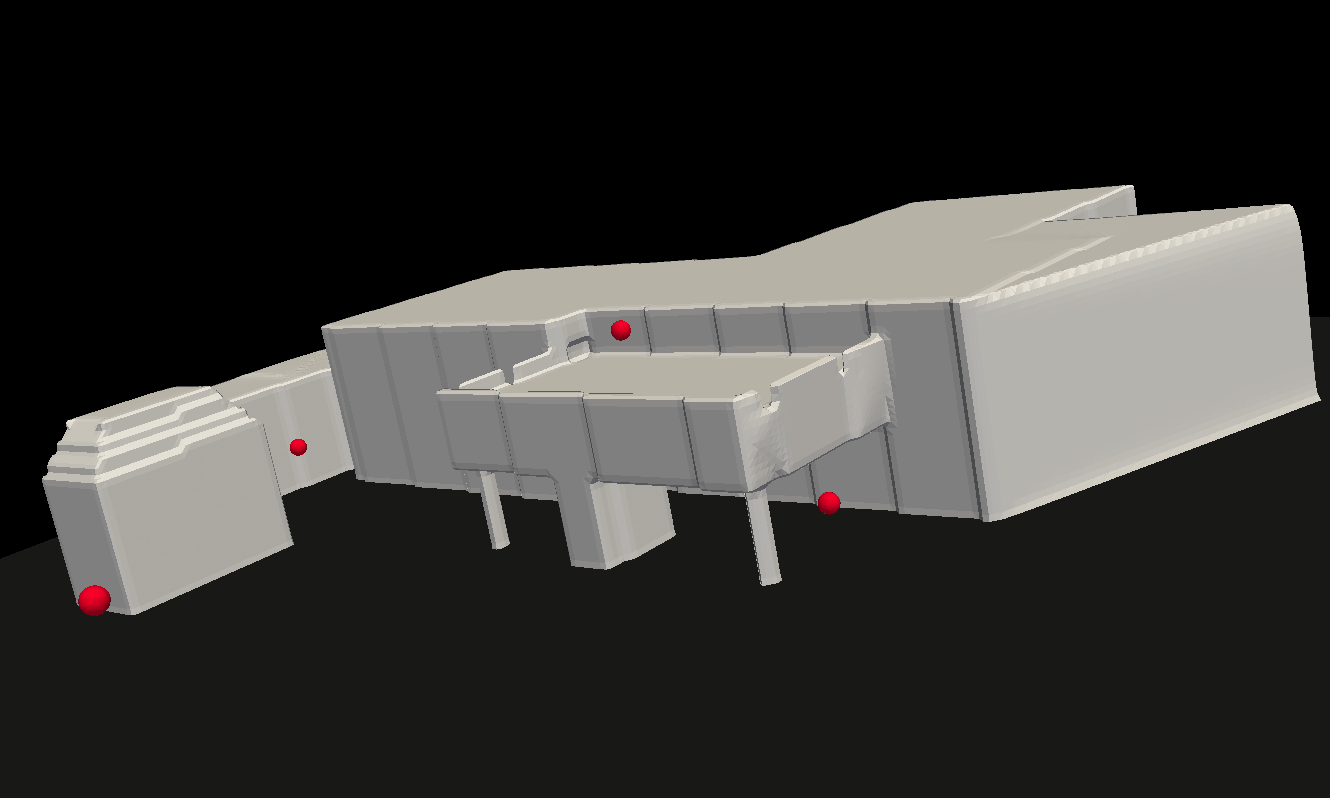}
      \includegraphics[width=5in]{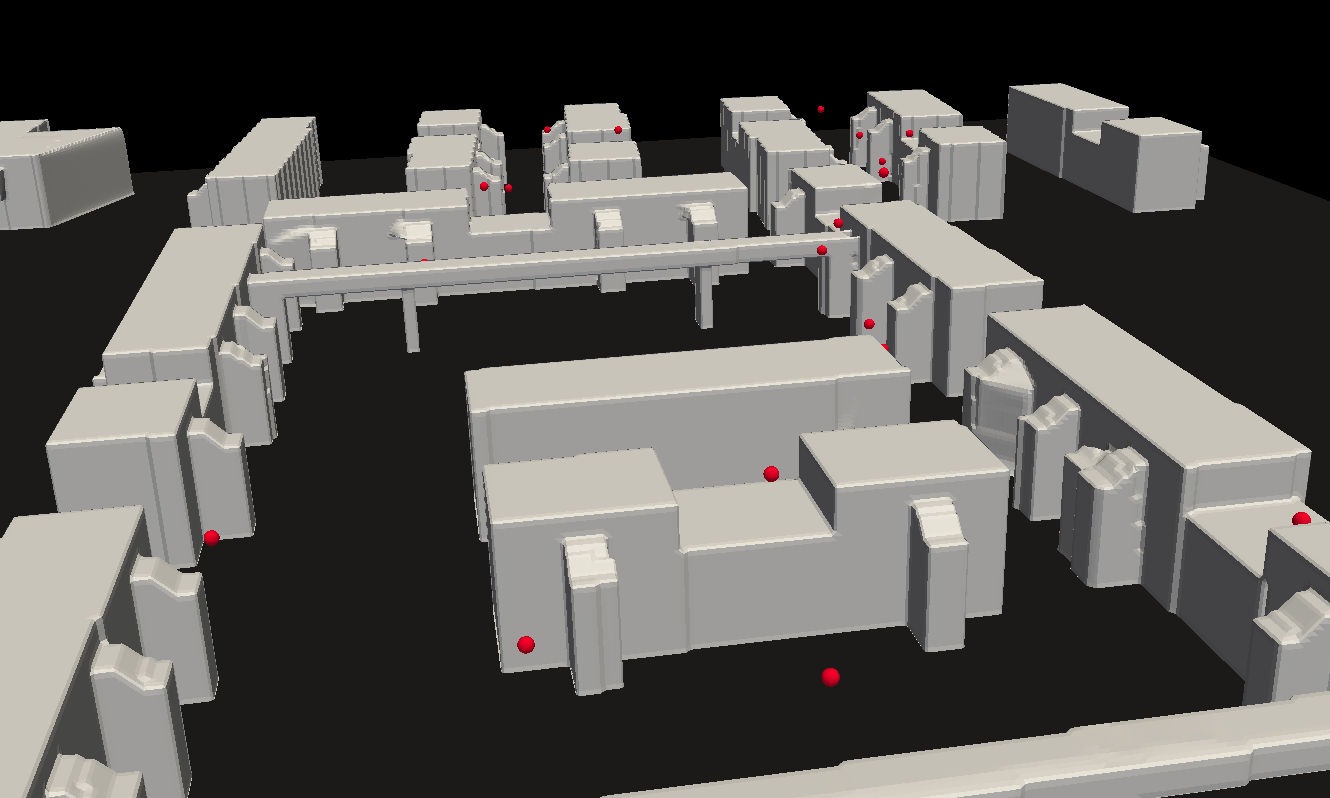}
      \caption{Snapshots demonstrating the exploration of an initially unknown
3D urban environment using sparse sensor measurements. The red spheres
indicate the vantage point. The gray surface is the reconstruction of the
environment based on line of sight measurements taken from the sequence of
vantage points. New vantage points are computed in virtually real-time using
{\bf 3D-CNN}.}

      \label{fig:3d-urban}
    \end{figure}

\begin{figure}[thpb]
      \centering
      \includegraphics[width=5in]{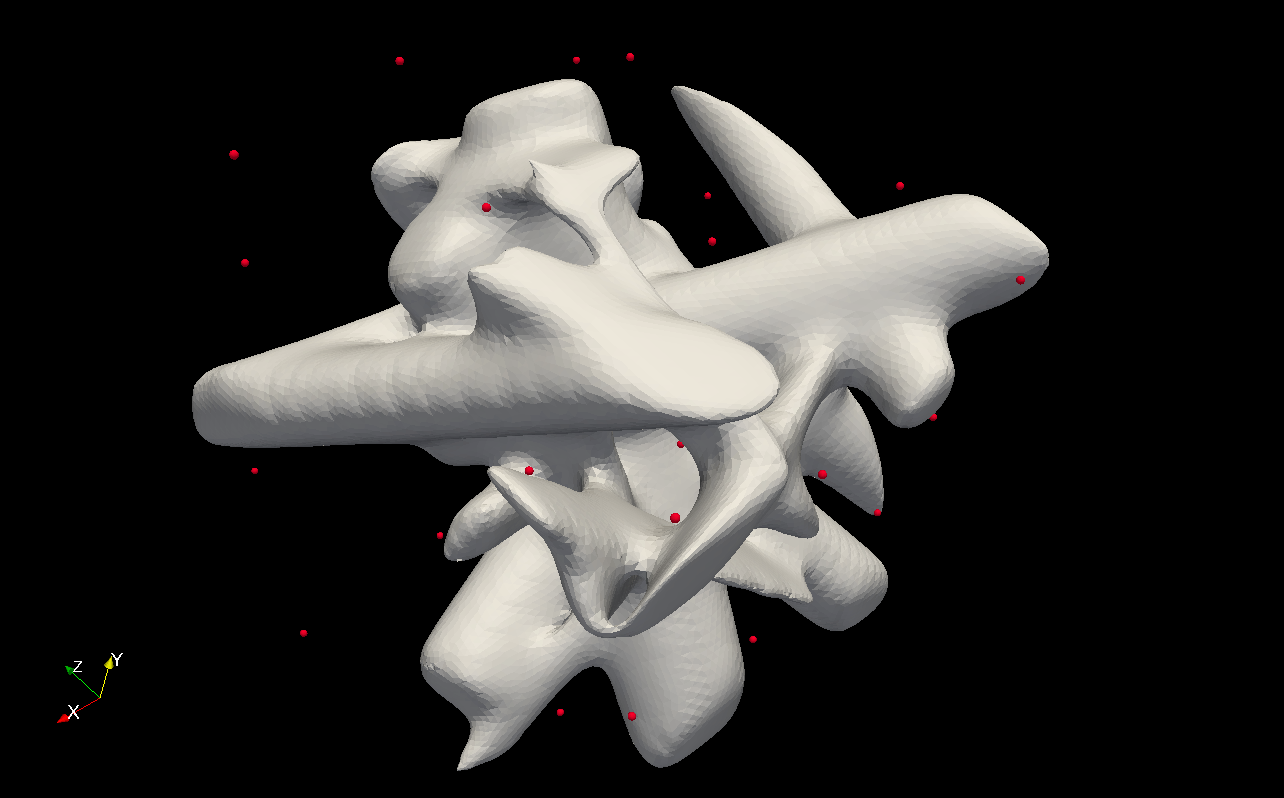}
      \includegraphics[width=5in]{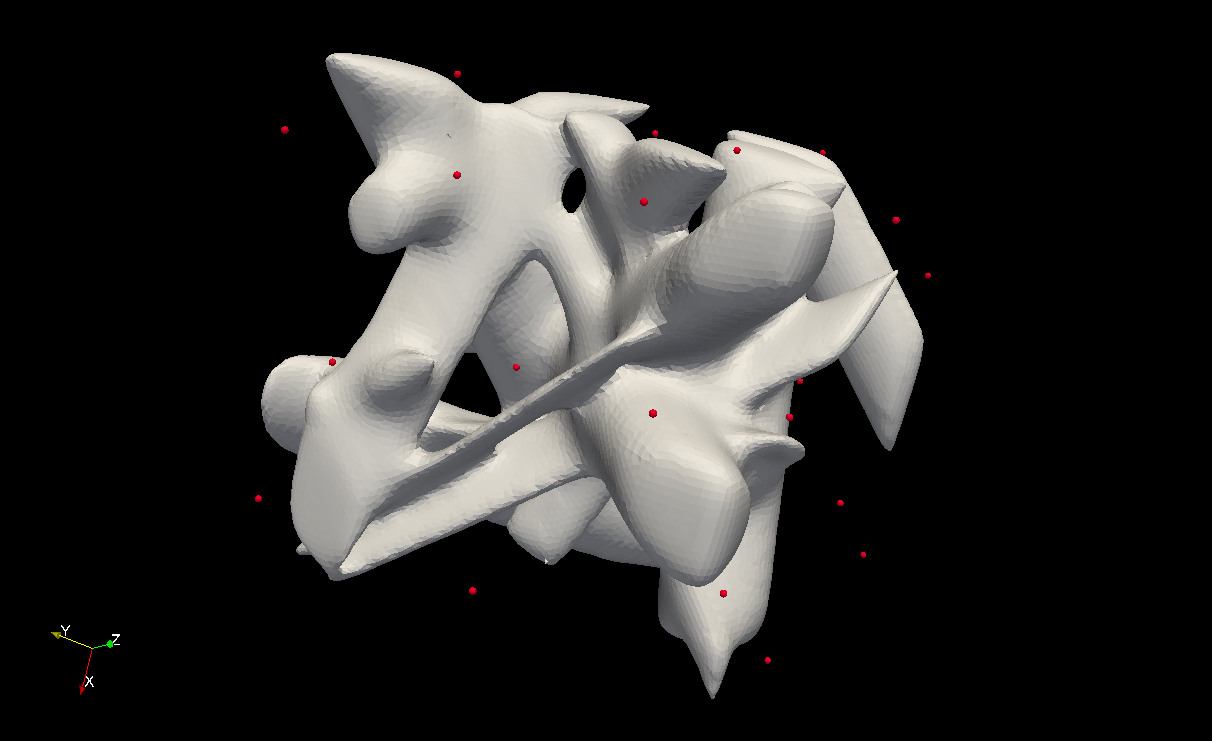}
      \caption{Snapshots of {\bf 3D-CNN} applied to exploration of a cluttered scene.}
      \label{fig:3d-pipes}
    \end{figure}

We present a 3D simulation of a 250m$\times$250m environment based on Castle Square Parks in Boston.
Figure \ref{fig:3d-urban} for snapshots of the algorithm in action.
The map is discretized as a level set function on a $768\times768\times64$ voxel grid.
At this resolution, small pillars are accurately reconstructed by our exploration algorithm.
Each step can be generated in 3 seconds using the GPU or 300 seconds using the
CPU. Parallelization of the distance function computation will further reduce the computation time significantly.
A map of this size was previously unfeasible. 
Lastly, Figure~\ref{fig:3d-pipes} shows snapshots from the exploration of a
more challenging, cluttered 3D scene with many nooks.

\section{Conclusion}
From the perspective of inverse problems, we proposed a greedy algorithm for
autonomous surveillance and exploration. We show that this formulation can be
well-approximated using convolutional neural networks, which learns
geometric priors for a large class of obstacles. The inclusion of shadow
boundaries, computed using the level set method, is crucial for the success of
the algorithm.
One of the advantages of using the gain function \eqref{gain-func}, an
integral quantity, is its stability with respect to noise in positioning and sensor
measurements. In practice, we envision that it can be used in
conjuction with SLAM algorithms \cite{durrant2006simultaneous,
bailey2006simultaneous} for a wide range of real-world applications.

One may also consider $n$-step greedy algorithms, where $n$ vantage points are
chosen simultaneously.  However, being more greedy is not necessarily better.
If the performance metric is the cardinality of the solution set, then it is
not clear that multi-step greedy algorithms lead to smaller solutions. We saw in
section~\ref{sec:greedy} that, even for the single circular obstacle, the greedy
surveillance algorithm may sometimes require more steps than the exploration
algorithm to attain complete coverage.

If the performance metric is based on the rate in which the objective function
increases, then a multi-step greedy approach would be appropriate.  However, on a grid
with $m$ nodes in $d$ dimensions, there are $\bigo(m^{nd})$ possible
combinations.  For each combination, computing the visibility and gain function
requires $\bigo(nm^d)$ cost. In total, the complexity is $\bigo(nm^{d(n+1)})$,
which is very expensive, even when used for offline training of a neural
network. In such cases, it is necessary to selectively sample only the relevant
combinations. One such way to do that, is through a tree search algorithm.


\section*{Acknowledgment}
This work was partially supported by NSF grant DMS-1913209.

%
%


\bibliographystyle{plain}  
\bibliography{references}        
\index{Bibliography@\emph{Bibliography}}



\end{document}